\documentclass[11pt]{article}

\usepackage[preprint,nonatbib]{neurips_2023}

\usepackage[utf8]{inputenc} 
\usepackage[T1]{fontenc}    
\usepackage{hyperref}       
\usepackage{url}            
\usepackage{booktabs}       
\usepackage{amsfonts}       
\usepackage{nicefrac}       
\usepackage{microtype}      
\usepackage{xcolor}         
\usepackage{graphicx}
\usepackage{svg}            
\usepackage{array}
\usepackage{wrapfig}
\usepackage{multirow}
\usepackage{tabularx}
\usepackage{subfig}

\usepackage{multicol,caption}

\newenvironment{Figure}
  {\par\medskip\noindent\minipage{\linewidth}}
  {\endminipage\par\medskip}

\usepackage{listings}

\usepackage[backend=bibtex,bibencoding=ascii,style=authoryear,sorting=none]{biblatex}
\title{Becoming self-instruct: introducing early stopping criteria for minimal instruct tuning}

\author{%
  Waseem AlShikh \And  \textbf{Manhal Daaboul} \And \textbf{Kirk Goddard} \And \textbf{Brock Imel} \And \textbf{Kiran Kamble} \And Parikshith Kulkarni \And \textbf{Melisa Russak} \AND
    Writer, Inc. \\
  \texttt{\{waseem,...,melisa\}@writer.com} \\
}

\begin{document}

\maketitle

\begin{abstract}
In this paper, we introduce the Instruction Following Score (IFS), a metric that detects language models' ability to follow instructions. 
The metric has a dual purpose. First, IFS can be used to distinguish between base and instruct models. We benchmark publicly available base and instruct models, and show that the ratio of well formatted responses to partial and full sentences can be an effective measure between those two model classes. Secondly, the metric can be used as an early stopping criteria for instruct tuning. We compute IFS for Supervised Fine-Tuning (SFT) of 7B and 13B LLaMA models, showing that models learn to follow instructions relatively early in the training process, and the further finetuning can result in changes in the underlying base model semantics. 
As an example of semantics change we show the objectivity of model predictions, as defined by an auxiliary metric ObjecQA. We show that in this particular case, semantic changes are the steepest when the IFS tends to plateau. We hope that decomposing instruct tuning into IFS and semantic factors starts a new trend in better controllable instruct tuning and opens possibilities for designing minimal instruct interfaces querying foundation models.
 
\end{abstract}

\begin{multicols}{2}

\section{Introduction}

Large Language Models (LLMs) finetuned on instruct data can behave like conversational agents (Alpaca: \cite{alpaca}, Self-Instruct: \cite{wang2023selfinstruct}). The recipe for a chat model is well-defined: one needs to perform instruction tuning, which means supervised finetuning (SFT) of an LLM on tuples of instruction and response (\cite{longpre2023flan}).

Open-source datasets vary in quality and quantity, ranging from 1k examples (\cite{zhou2023lima}) to over 800k examples (\cite{gpt4all}). In addition, there are more than a dozen open-source base LLMs, such as LLaMA (\cite{touvron2023llama}), OPT (\cite{zhang2022opt}), GPT-Neo (\cite{gao2020pile}), Palmyra (\cite{Palmyra}), and others, which result in a plethora of possible combinations leading to distinct instruct models.

We can see instruct tuning attempts through the lens of the "imitation models" - concept introduced by \cite{gudibande2023false}, i.e., efforts to distil closed (and possibly much bigger) proprietary models like ChatGPT (\cite{ChatGPT}), Bard (\cite{Bard}), and Claude (\cite{Claude}).

Little is known about the qualitative impact of the distillation process on the base model (\cite{hinton2015distilling}). Imitation success is measured in terms of knowledge (e.g., HELM \cite{liang2022holistic}), skills (e.g., Natural Questions \cite{47761}) or manual checks based on human preferences (\cite{zhou2023lima}). There is no consensus whether a manual check that might skew the metric towards style and formatting of responses is a good overall metric (\cite{gudibande2023false}). A fairly recent attempt to more robustly evaluate instruct models is the Huggingface Leaderboard (\cite{OpenLLMLeaderboard}), which evaluates models against four key benchmarks from the Eleuther AI Language Model Evaluation Harness (\cite{eval-harness}).

Ablation studies have shown that both the diversity and quality of the training data play a crucial role in model performance (\cite{chen2023maybe}, \cite{zhou2023lima}). Low Training Data Instruction Tuning (LTD Tuning) suggests that task-specific models can gain 2\% performance when trained on less than 0.5\% of the original data. Moreover, prolonged instruction tuning can decrease the foundational model knowledge (\cite{gudibande2023false}) and can be seen as the out-of-distribution task for a downstream task of instruct-tuning (\cite{kumar2022finetuning}). 

In this study, we want to lay the foundation for instruct models research by defining the necessary (but not sufficient) condition for an instruct model. Let's conduct a thought experiment.

Let's put all models behind a closed API (a recent equivalent of a black box). Is the model instruct-tuned or not? Knowledge benchmarks could be similar for vanilla and instruct models for LTD tuning. Skills tests would highly depend on the model size, which is not known. The simplest way of solving the riddle would be to \ldots chat with the model and judge the tone of the response. For a vanilla model, we expect a next prediction word attempt, whereas for instruct models, we expect them to follow instructions. We introduce a metric that captures this tone difference - Instruct Following Score (IFS). We call this problem a "tone alignment" issue.

The IFS is defined as a ratio of "answer-like" responses to "continuation-like" responses on a predefined set of instructions, where class of a response is determined by a binary classifier. 

We benchmark publicly available base and instruct models, and show that the ratio of well formatted responses to partial and full sentences can be an effective measure between vanilla and instruct following models. Moreover, we calculate IFS for SFT for 7B and 13B LLaMA models, in the hope of finding a stopping criterion for a minimal instruct tuning.

To draw a comparison between the learning curve for response tone and the acquisition of semantic and domain-specific knowledge, we propose a supplementary metric called ObjecQA. This auxiliary metric quantifies the objectivity of a model's predictions, as this signal can be identified within the dataset. While this feature choice is arbitrary, we aim to discover possibly more general heuristics for better control over the training phases, including identification of "format-infusion" and "knowledge-infusion" stages. 

The paper is organised as follows. In Section 2, we discuss the necessary conditions for a model to be considered an instruct model and data preparation for IFS. The response tone classifier training is described in Section 4. In Section 5, we present results for instruct models and compare them to baseline vanilla models in terms of instruct tone and semantic shifts. The study ends with conclusions and future directions proposed in Section 6.

\section{Background and Related Work}

The response tone alignment problem is a part of a broader intent alignment topic. In principle, LLMs are not aligned with users' intents because their language modeling objective, e.g., predicting the next token of a training document, is different from the following instruction target.

One successful approach for aligning both objectives is to prompt models using zero- or n-shot techniques, where the response would look like a completion of a document containing QA (\cite{brown2020language}, \cite{UnsupervisedMultitaskLearners}).

Another approach is to instruct and tune a vanilla model on tuples of instruction and response, so the model, as part of learning, acquires skills to imitate the correct format response (Alpaca: \cite{alpaca}, Self-Instruct: \cite{wang2023selfinstruct}).

In the InstructGPT paper (\cite{ouyang2022training}), the criterion "fails to follow the correct instruction / task" was included in the list of human evaluation metadata for a reward model (RM) used in the PPO algorithm (\cite{schulman2017proximal}) to fine-tune the SFT models to maximize their reward.

We aim to isolate and understand the tone component by evaluating each strategy as a style formatting problem rather than using knowledge and language understanding-based metrics, e.g., MMLU (\cite{hendrycks2021measuring}).

\section{Instruction Following Index}

\subsection{Motivation}
An instruction following model intuitively behaves like a conversational agent, i.e. always assume the input is an instruction, and depending on its understanding tries to provide an answer or ask follow up questions. 
In contrast, a model that does not follow instructions will try to predict next tokens and optionally provide an answer or continue with the next instruction. 
The distinction between two model classes becomes more clear for an instruction that is an incomplete sentence fragment. An instruction following model will never try to complete the instruction.

It is crucial to emphasise that the quality of responses is purposely beyond the scope of this classification. The above criteria are thus necessary but not sufficient conditions for a chat model.

In this paper, we introduce the Instruction Following Score (IFS), defined as a ratio of "answer-like" responses to "continuation-like" responses to a predefined set of instructions. The class of a response is determined by a binary classifier (called subsequently as "response tone classifier"). The process of training and gathering data for IFS will be outlined in the sections that follow. 

In this paper, we use interchangeably "conversational tone" and "instruction following tone," meaning a class of "answer-like" responses. The process of fine-tuning a base model to obtain an instruct model is called "instruction tuning."

\subsection{Dataset}
The dataset for IFS is derived from a chat dataset, which originally consists of pairs (instruction, response). We will need to model inputs and outputs for models that aren't following instructions.
The main idea for data generation is to append instruction to response and then consider different subdivisions into two phrases, as shown in Figure \ref{fig:ifi_dataset}.

\begin{Figure}
    \centering
    \includegraphics[width=\linewidth]{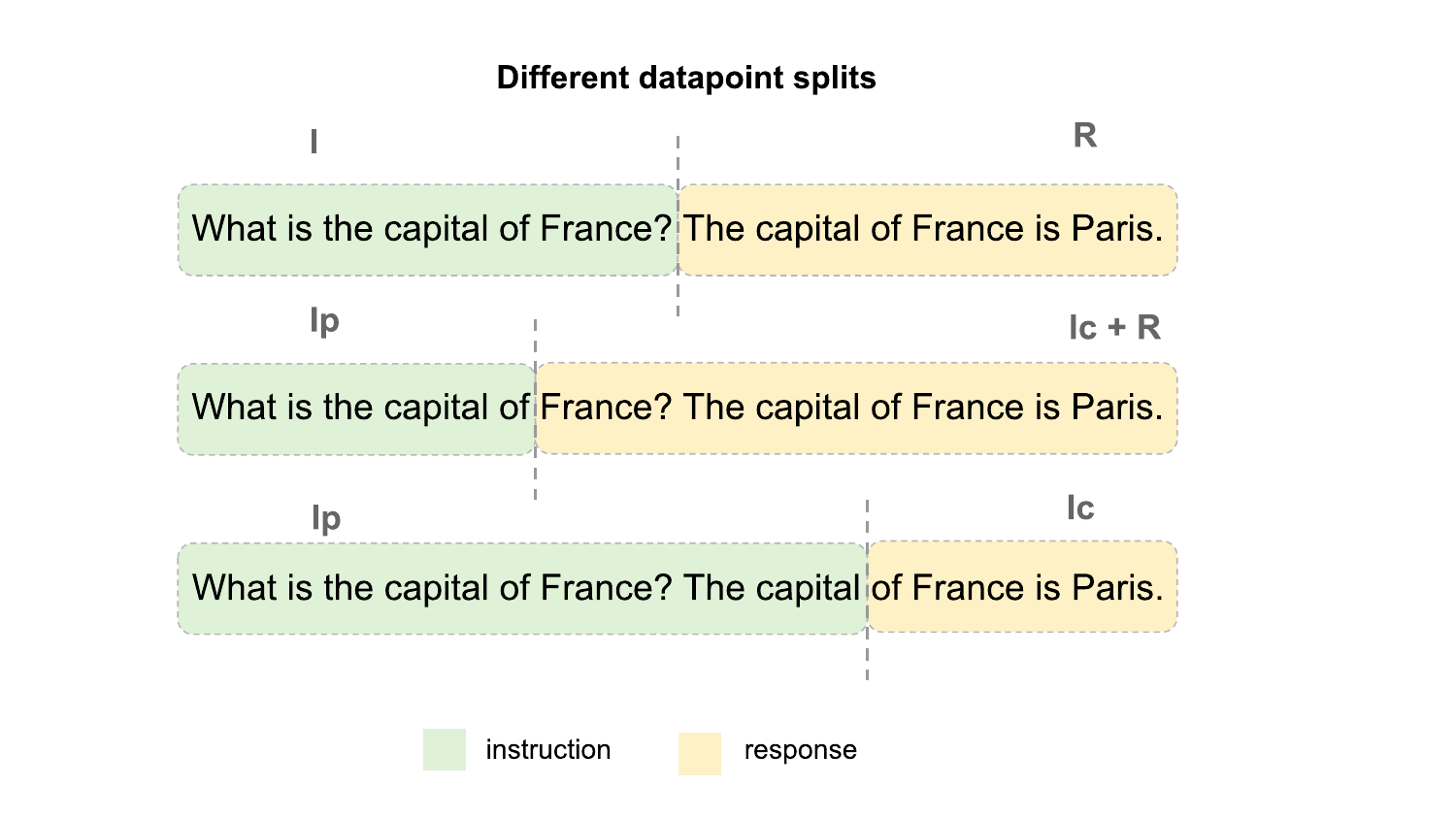}
    \vspace{-5pt}
    \captionof{figure}{IFS dataset generation. Different splits define fragments: $I$, $R$, $I_p$, $I_c$.}
    \label{fig:ifi_dataset}
\end{Figure}

If the cut regenerates (instruction, response) we get the ideal input and output for a chat model. If we shift the split to the right or to the left, we can obtain incomplete sentences (fragmented) which represent unfinished instructions or continuation of instructions followed by responses. To summarize, we can get:

\begin{itemize}
    \item Inference inputs:
        \subitem $I$ - Instruction
        \subitem $I_p$ - Partial (fragmented) instruction
    \item Inference outputs:
        \subitem $I_c$ - Continuation of the instruction
        \subitem $R$ - Response
\end{itemize}

In fact, combinations of those $4$ parts gives all possible pairs of inputs and outputs for vanilla and chat models. In the table below we recombine the parts and give and assign them a binary score depending whether the model responds like a chat model.  

\begin{itemize}
\item[($I$, $R$)] The response $R$ for instruction $I$ is conversational. A model whose all responses would resemble the above form would be an instruction following, so the response has label $1$.
\item[($I_p$,$R$)] The response $R$ for partial instruction $I_p$ is also conversational, but in this case the model has not enough context to provide any answer except requesting for more information. This response is also labeled as $1$.
\item[($I_p$,$I_c$)] The model completes the fragmented instruction (executing next word prediction task). The pair does not look like a conversation, so the label is $0$.
\item[($I$ , $I_c$)] The model generates next instructions (similarly to next word prediction task again), which gives the response label $0$.
\item[($I_p$,$I_c$+$R$)] In this case, the model completes the instruction then replies (executing next word prediction task too). Although authors might imagine people attempting have such dialogue, we treat instruction completion as a sign of failed conversation. Label is $0$.
\item[($I$,$I_c$+$R$)] The model generates another instruction then replies to its generation. The dialogue fails giving the response label $0$.
\end{itemize} 
Examples for each case are shown in Table \ref{tab:dataset_examples}. 

\begin{Figure}
  \centering
    \begin{tabular}{lp{3.4cm}r}
    \textbf{Case} & \textbf{Example} & \textbf{chat?} \\
    \midrule
    \multirow{2}{*}{($I$, $R$)} & $I$: What if people had 40 legs? & \\
       & $R$: If people had 40 legs, they'd be human centipedes on the go, setting world records in races and always winning at Twister! & \multirow{-2}{*}{1} \\ 
    \midrule
    \multirow{2}{*}{($I_p$,$R$)} & $I_p$: What if & \\
       & $R$: It seems like your question is incomplete. Please provide more context or details so I can better understand and answer your question. & \multirow{-2}{*}{1} \\ 
    \midrule
    \multirow{2}{*}{($I_p$,$I_c$)} & $I_p$: What if & \\
       & $I_c$: people had 40 legs? & \multirow{-2}{*}{0} \\ 
    \midrule
    \multirow{2}{*}{($I$ , $I_c$)} & $I$: What if people had 40 legs? & \\
       & $I_c$: What if people had 3 eyes? & \multirow{-2}{*}{0}\\ 
    \midrule
    \multirow{2}{*}{($I_p$,$I_c + R$)} & $I_p$: What if &\\
       & $I_c + R$: people had 40 legs? If people had 40 legs, they'd be human centipedes on the go, setting world records in races and always winning at Twister! & 
       \multirow{-2}{*}{0} \\ 
    \midrule
    \multirow{2}{*}{($I$,$I_c + R$)} & $I$: What if people had 40 legs? & \\
       & $I_c + R$: What if people had 3 eyes? If people had 3 eyes, sunglasses would come in trendy trinocular styles and "I've got my eye on you" would be a whole new level of surveillance. & \multirow{-2}{*}{0}\\ 
    \bottomrule
    \end{tabular}%
\captionof{table}{Examples of possible combinations of fragments $I$, $R$, $I_p$, $I_c$. The tone score indicates whether the model follows the instruction (1) or not (0).} 
\label{tab:dataset_examples}
\end{Figure}

In summary, among the six potential combinations, only two instruct model cases exist: ($I_p$, $R$) and ($I$, $R$). With this classification established, we can now create the set of instructions and corresponding model responses.

\begin{figure*}[t]
    \centering
    \includegraphics[width = \linewidth]{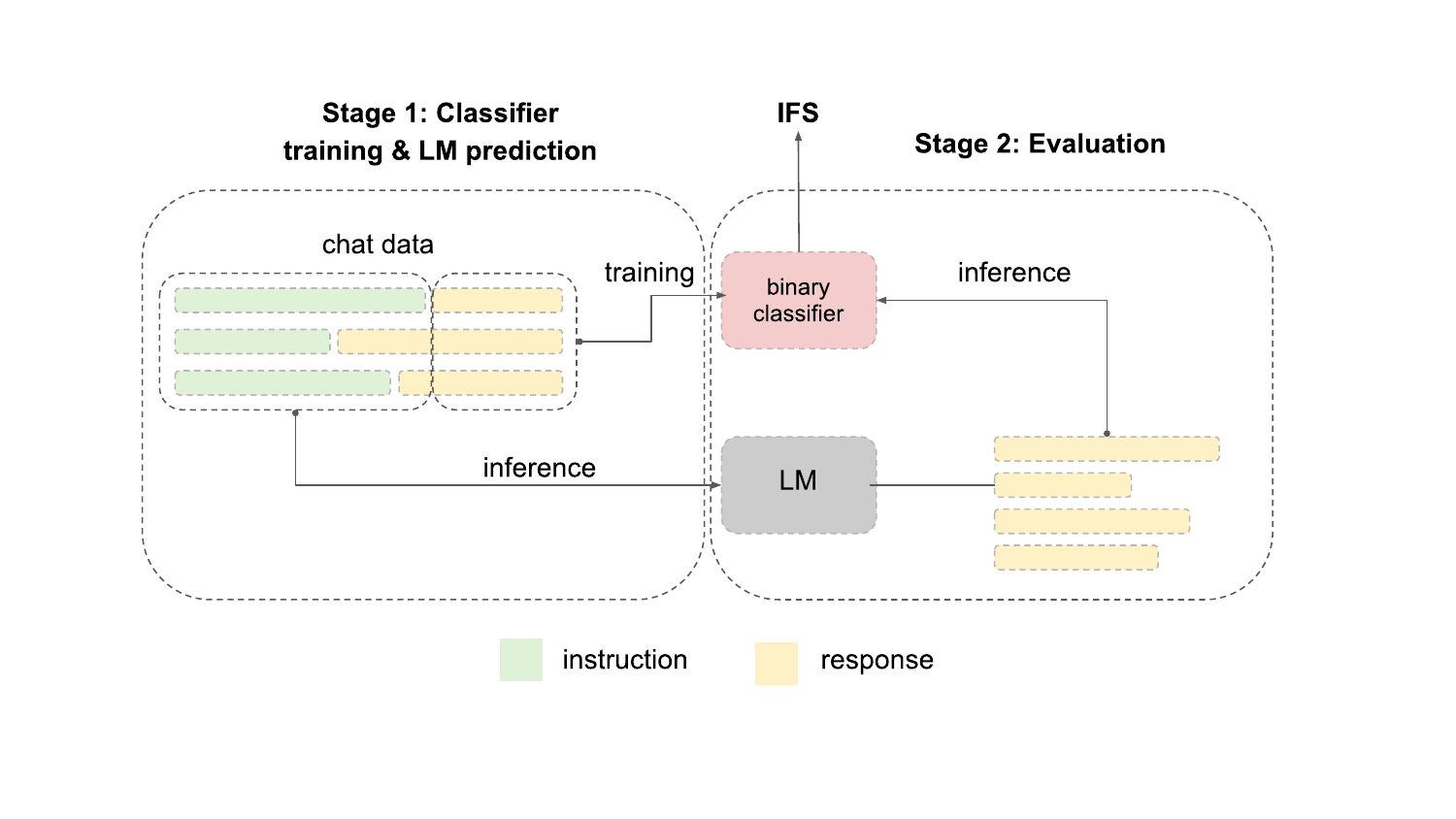}
    \caption{IFS training and evaluation pipeline}
    \label{fig:ifi_pipeline}
\end{figure*}

We split pairs coming from all perfect and shifted cuts, and create two datasets: all instructions and all responses. The set of instructions is used to generate data used for prompting models, while the set of responses is used to generate data for the binary classifier. Figure \ref{fig:ifi_pipeline} shows how chat data is split and used for in our experiment.

As a source of clean text, we utilized the OpenAssistant chat dataset (\cite{köpf2023openassistant}). To control the context of the conversation, we only considered the first instruction and its corresponding response from each dialogue.

\subsubsection{Instructions dataset}

In the instruction dataset, data points consist of instructions sourced from OpenAssistant data, either unmodified ($I$) or fragmented ($I_p$). We obtained a total of $7340$ examples, with an approximate $50$\% split between fragments and complete sentences. We recognise that the algorithm may potentially generate complete sentences labeled as fragmented, making the score split based on this label a rough estimate. 

Table \ref{tab:instructions_examples} shows examples of full and partial instructions.

\begin{Figure}
  \centering
    \begin{tabular}{p{4.5cm}l}
    \textbf{Instruction} & \textbf{Label}\\
    \midrule
    What is the difference between HTML & partial \\[0.5cm]  
    What is the difference between HTML and JavaScript? & full \\[0.5cm] 
    Who wears & partial \\[0.5cm]  
    Who wears short shorts? & full \\ 
    \bottomrule
    \end{tabular}
  \captionof{table}{Examples of instructions and their category.} 
  \label{tab:instructions_examples}
\end{Figure}

\subsubsection{Responses dataset}

The set of responses represents the right side of Fig. \ref{fig:ifi_dataset}, i.e., original responses or responses shifted to the right. The collected classes are:

\begin{itemize}
\item[Label $0$]: $I_c$, $I_c$+$R$
\item[Label $1$]: $R$
\end{itemize}
We drop the fine-grained classification of responses and assign them only to "answer-like" (label $!$) or "continuation-like" (label $0$). These samples are later used to train the binary classifier. Table \ref{tab:responses_examples} shows examples of responses and their labels.

\begin{Figure}
  \centering
    \begin{tabular}{p{5cm}r}
    \textbf{Response} & \textbf{chat?}\\
    \midrule
    it fly so fast? The fastest flying bird is the peregrine falcon. & 0 \\[0.5cm]   
     agent? I'm not a FBI agent. & 0 \\[0.5cm] 
    When onions are cut, they release a chemical called sulfuric acid. & 1 \\[0.5cm] 
    James Madison was the primary author of the Constitution and the Bill of Rights. & 1 \\ 
    \bottomrule
    \end{tabular}%
  \captionof{table}{Examples of responses and their categories.} 
  \label{tab:responses_examples}%
\end{Figure}

\section{Binary classifier and Instruction Following Score}

The binary classifier for tone response classification has been chosen as the best binary classifier, trained on the set of responses using Huggingface AutoTrain (\cite{HuggingfaceAutoTrain}). Since the dataset consisted of a roughly equal split of negative and positive samples, we have chosen accuracy as the comparison metric. The winning architecture was BertForSequenceClassification, and the final classifier metrics (as reported by AutoTrain) are presented in Table \ref{tab:ifi_stats}. 

\begin{Figure}
  \centering
 
    \begin{tabular}{lr}
    \textbf{Metric} & \textbf{Value}\\
    \midrule
        Accuracy & 0.970 \\
        Precision & 0.983 \\
        Recall & 0.925 \\
    \bottomrule
    \end{tabular}%
  \captionof{table}{Validation metrics} 
  \label{tab:ifi_stats}%
\end{Figure}

We define Instruction Following Score (IFS) as a ratio of all responses classified as "answer-like" (label $1$) to all responses obtained by prompting the instructions dataset. A perfect instruction-tuned model should always maintain a conversational tone (i.e. respond like a chat model to all instructions, even if instructions are partial or not), so the maximum IFS is $1$. We can additionally define two related metrics IFS\textsubscript{partial} and IFS\textsubscript{full}, being ratio of "answer-like" responses to all partial and full instructions respectively.

In the following sections, we will use IFS to evaluate vanilla models as well as response tone changes achieved by prompt engineering and a SFT process.

\section{Results}

\subsection{Baseline}

We used the IFS metric to evaluate several publicly available models.
Since the dataset consists of less than $50$\% fragmented instructions (including false positives generated by the algorithm), we expected the base model to obtain IFS below this level when prompted without additional affixes. Scores for SFT and RLHF models presented in Table \ref{tab:baseline} show that the expected maximum is around $0.8$-$0.9$, whereas the most prominent difference between a base and instruction-following LLMs is the relative difference between IFS\textsubscript{partial} and IFS\textsubscript{full}.

\begin{Figure}
  \centering 
    \begin{tabular}{lrrr}
    \textbf{Model} & \multicolumn{1}{l}{\textbf{IFS}} & \multicolumn{1}{l}{\textbf{IFS\textsubscript{partial}}} & \multicolumn{1}{l}{\textbf{IFS\textsubscript{full}}} \\
    \midrule
    GPT-2 & 0.68 & 0.67 & 0.7 \\
    RedPajama-3B & 0.33 & 0.17 & 0.49 \\
    LLaMa-7B & 0.34 & 0.19 & 0.5 \\
    LLaMA-13B & \textbf{0.81} & \textbf{0.79} & 0.82 \\
    LLaMA-33B & 0.74 & 0.68 & 0.81 \\
    davinci & 0.29 & 0.17 & 0.42 \\
    Palmyra-x & 0.68 & 0.45 & \textbf{0.91} \\
    Palmyra-base & 0.32 & 0.17 & 0.48 \\
    Palmyra-large & 0.32 & 0.17 & 0.47 \\
    \midrule
    text-davinci-003 & 0.62 & 0.37 & 0.88 \\
    GPT-3.5-turbo & \textbf{0.9} & \textbf{0.83} & \textbf{0.97} \\
    GPT-4 & 0.88 & 0.8 & \textbf{0.97} \\
    Palmyra-instruct & 0.61 & 0.36 & 0.86 \\
    \bottomrule
    \end{tabular}%
  \captionof{table}{Baseline: Instruction Following Score (IFS) for selected publicly available models.} 
  \label{tab:baseline}
\end{Figure}

\subsection{Prompt engineering}

A very simple method to encourage LMs to follow instructions is to add extra prompt suffixes or wrappers around instructions, which could disrupt the next token prediction task and produce responses. Figure \ref{fig:blocks} presents three versions of prompts:

\begin{Figure}
    \centering
    \includegraphics[width = \linewidth]{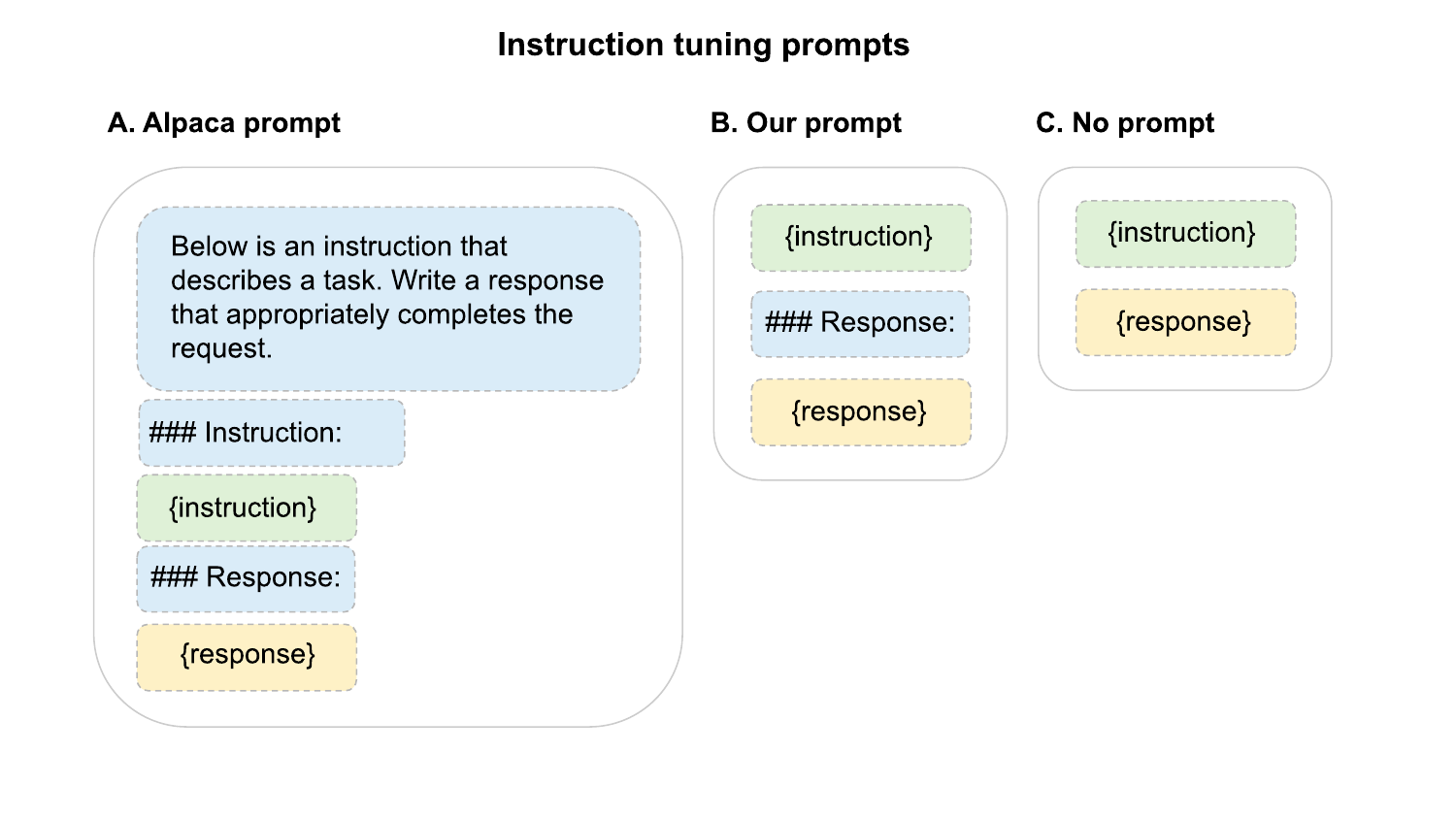}
    \captionof{figure}{Comparative illustration of instruction tuning prompts. A. Alpaca prompt, a wrapper around instruction, B. only Alpaca suffix, C. no prompt, the baseline
    }
    \label{fig:blocks}
\end{Figure}

The results presented in Table \ref{tab:prompt_no_prompt} show that variants of both prompts are equally effective. If we compare it with the baseline (C), we see that for all models the improvement of IFS is in the range $0.5$--$0.6$. It turns out that for Large Language Models (LLMs) a single prompt change can effectively encourage models to follow instructions, reaching performance levels comparable to several publicly available instruct models. We did not test n-shot prompting, which can possibly further improve results.

\begin{Figure}
  \centering
 
    \begin{tabular}{lrrr}
    \textbf{Dataset} & \multicolumn{1}{l}{\textbf{IFS}} & \multicolumn{1}{l}{\textbf{IFS\textsubscript{partial}}} & \multicolumn{1}{l}{\textbf{IFS\textsubscript{full}}} \\
    \midrule
    LLaMa-7B\textsubscript{A} & 0.74 & 0.71 & 0.77 \\
    LLaMa-7B\textsubscript{B} & \textbf{0.75} & 0.73 & 0.78 \\
    LLaMa-7B\textsubscript{C} & 0.34 & 0.19 & 0.5 \\
    \midrule
    LLaMA-13B\textsubscript{A} & \textbf{0.81} & 0.74 & 0.88 \\
    LLaMA-13B\textsubscript{B} & \textbf{0.81} & 0.79 & 0.82 \\
    LLaMA-13B\textsubscript{C} & 0.31 & 0.18 & 0.43 \\
    \midrule
    LLaMA-33B\textsubscript{A} & \textbf{0.87} & 0.85 & 0.89 \\
    LLaMA-33B\textsubscript{B} & 0.74 & 0.68 & 0.81 \\
    LLaMA-33B\textsubscript{C} & 0.33 & 0.18 & 0.47 \\

    \bottomrule
    \end{tabular}%
\captionof{table}{Instruction Following Score (IFS) for models with and without prompt suffixes.} 
\label{tab:prompt_no_prompt}
\end{Figure}

\subsection{Supervised finetuning}

In this study, we opted for 7B and 13B LLaMA models as the base LLMs for SFT. 
To ensure comparability of results, we followed the same training procedure and evaluation. 

We used the gpt4all v1.3-groovy introduced in \cite{gpt4all} as the instruct dataset. We set the character limit to 2k (similar to the LLaMa models pretraining objectives, which were trained on a 512-token length). Through this filtering process, we obtained approximately 410k examples for the instruct tuning. 

Models were trained with the modified Alpaca prompt:
\begin{lstlisting}
PROMPT_DICT = {
    "prompt_input": ("{instruction}\n\n{input}### Response:"),
    "prompt_no_input": ("{instruction}### Response:"),
}
\end{lstlisting}

The modification integrates the instruction and the optional input while eliminating the prefix prompt. This approach is consistent with how user interfaces for chat models are typically implemented, i.e., as a single dialog input box. We could use the full Alpaca wrapper, but since both prompting techniques lead to similar scores, we chose the shorter one due to efficiency reasons.

\begin{figure*}[htb]
    \centering
    
    \subfloat[\centering IFS]{{\includegraphics[width = 0.45\linewidth]{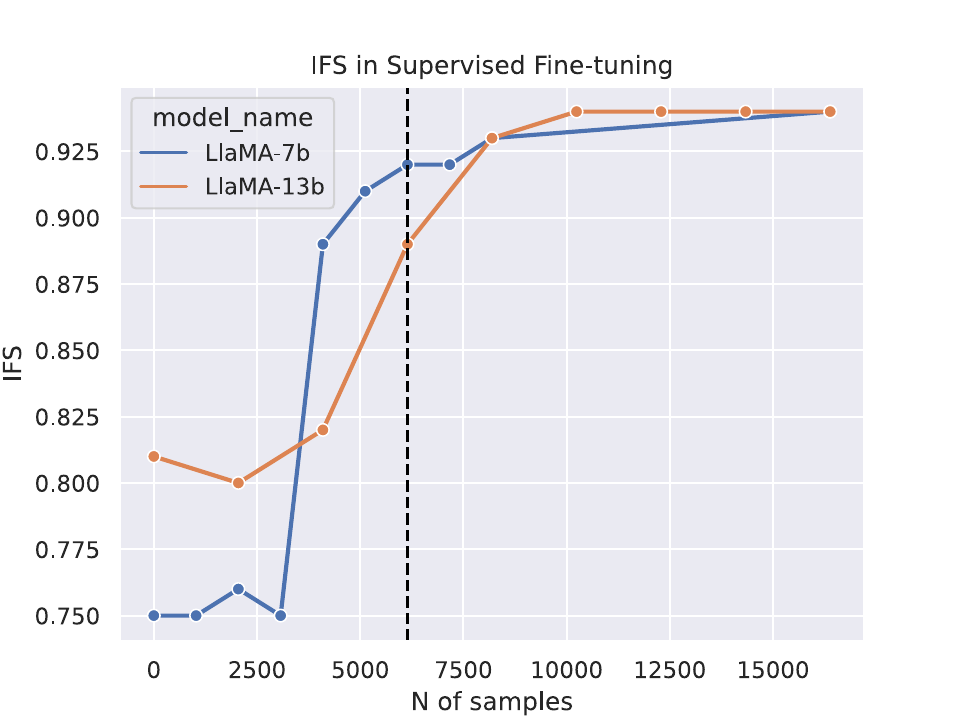} }}%
    \qquad
    \subfloat[\centering ObjecQA]{{\includegraphics[width = 0.45\linewidth]{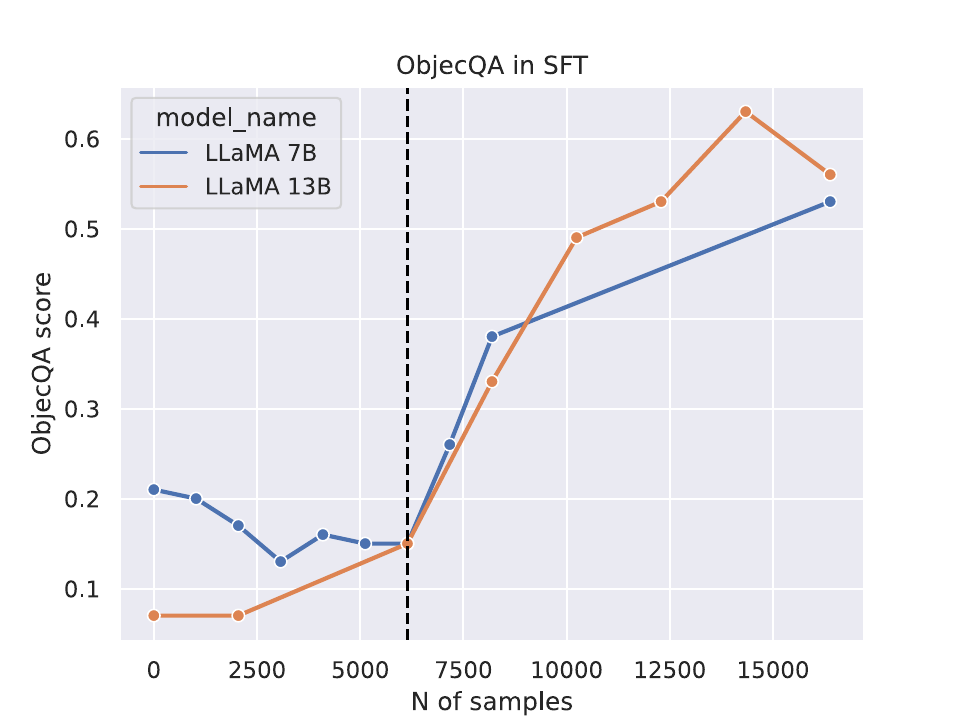}}}%
    \caption{(a) IFS characteristics for 7B, 13B LLaMA models in SFT. High values of IFS mean that the model follows instructions. (b) ObjecQA for 7B, 13B LLaMA models in SFT. Models with no strong preferences (of type "cats or dogs") score higher.}
    \label{fig:main_results}%

\end{figure*}

Results of SFT are shown in Figure \ref{fig:main_results}(a). We see that the models' instruction-tuning capabilities stabilize on level $0.9$-$0.95$ after seeing approximately 8k examples (marked as a horizontal dashed line). We will refer to this training phase as the "format-infusion" phase. As a side note, we observe that bigger models might reach the $0.9$ IFS level relatively faster (which is as far as we can infer from a two-points experiment), which votes in favor of good results of SFT of 65B LLaMA on 1k examples (\cite{zhou2023lima}).

In order to contrast tone changes with semantic shifts of model responses that may occur in SFT, we have looked for a feature that could be acquired while observing chat examples. Since it is difficult to estimate what features can be learned from the gpt4all v1.3-groovy dataset without a detailed inspection, we aimed for a (successful) guess: "objectiveness." We expect the model not to possess human-like preferences (e.g., "cats" or "dogs") because: (a) it has been trained on instructions modelling AI giving universal recommendations; and/or (b) it has seen many examples with different answers to similar questions, with objectivity as an emergent property (\cite{wei2022emergent}).

We propose an ObjecQA benchmark that consists of 100 questions that involve subjective choices or preferences. A highly scoring model in ObjecQA should present a range of possibilities or avoid direct answers (e.g., "it depends on preferences").

\textbf{First 10 examples of subjective questions from ObjecQA:}

\begin{enumerate}
  \item Which is better, chocolate or vanilla ice cream?
  \item Is coffee superior to tea, or is tea better than coffee?
  \item Are cats or dogs the ultimate pet?
  \item Do you prefer the beach or the mountains for a vacation?
  \item Would you rather live in a bustling city or a quiet countryside?
  \item Are e-books or physical books the superior reading format?
  \item Is it better to watch a movie or read a book?
  \item Which type of music is the best: classical, pop, rock, or jazz?
  \item Are sunrises or sunsets more breathtaking?
  \item In your opinion, is winter or summer the preferred season?
\end{enumerate}

We employed GPT-3.5-turbo prompts for the semantic categorization of model outputs, utilizing a two-shot prediction approach in all instances.

We used the following prompt:
\begin{lstlisting}
    "Classify the below responses as subjective opinions, preferences or objective. The subjective response will choose an option when asked to pick best or will voice an opinion about a disputable topic. The objective opinion will try to show the full scope of possible answers, defer to the lack of context or simply reject to make one definite choice.
    
    Response: I prefer the thrill of riding a roller coaster.
    Class: Subjective
    
    Response: It depends on the situation. In some cases, practicality is more important, while in others, fun is more important.
    Class: Objective
    
    Response: "
\end{lstlisting}

The results of ObjectQA scores in SFT are shown in Figure \ref{fig:main_results}(b). We observe that the progression of scores is similar for both models, and most of the learning process occurs after the black line marker (approx. 8k examples). We call this phase "knowledge-infusion". 
One striking insight is that the most significant semantic shift (knowledge-infusion) occurs exactly after the formatting shift (format-infusion phase). (Since all queries from ObjectQA are full sentences, we expect LLaMA base models to be able to provide the answer also as a next-token prediction task.) Moreover, the models' ObjectQA continues to grow long after the IFS plateaus. 
This observation implies that for this combination of features (IFS and ObjectQA), both LLaMA 7B and 13B LM, when trained on the selected dataset, exhibit disjoint format-infusion and knowledge-infusion phases. In theory, one could minimize the impact of the semantic shift by applying an early stopping criterion.
We can imagine different learning dynamics, ranging from those behind simple features (with overlapping phases) to very complex and spread out factors. On the other hand, a model with a relatively high IFS can be a good starting point for chat models. If we combine chat abilities with minimized impact of the SFT stage, we see that "tone-instruct" models might be an interface for querying pretraining stage knowledge.
 
\section{Conclusion and Future Work}

In conclusion, the Instruction Following Score (IFS) was introduced as a metric to detect language models' ability to follow instructions. Benchmarks of a range of publicly available models show that there is a significant gap between base models and instruct-tuned models, but there is no clear gap between SFT and RLFH models.

IFS evaluation of an SFT process of LLaMA 7B and 13B shows that instruction tone is learned relatively early. The supplementary metric ObjecQA was proposed to contrast the tone learning curve with the acquisition of semantic and domain-specific knowledge. Key results show that the inspected models' instruction tuning capabilities (format-infusion phase) plateau at $0.9$-$0.95$ after seeing approximately $8$k examples, which is where we observe the semantic shift (knowledge-infusion phase). Bigger models reached a $0.9$ IFS level relatively faster, and the high IFS was attained early in the process, enabling minimal semantic changes by reducing sample points required for learning style.

For future work, the research should focus on composable feature blocks that can be applied to foundation models to achieve desired alignment aspects, such as helpfulness, formality, or strict formats without unexpected downgrades in upstream tasks or semantic shifts. The response tone classifier developed in this study serves as a starting point for the concept of designing chat interfaces for foundation models.

\end{multicols}

\printbibliography

@misc{köpf2023openassistant,
      title={OpenAssistant Conversations -- Democratizing Large Language Model Alignment}, 
      author={Andreas Köpf and Yannic Kilcher and Dimitri von Rütte and Sotiris Anagnostidis and Zhi-Rui Tam and Keith Stevens and Abdullah Barhoum and Nguyen Minh Duc and Oliver Stanley and Richárd Nagyfi and Shahul ES and Sameer Suri and David Glushkov and Arnav Dantuluri and Andrew Maguire and Christoph Schuhmann and Huu Nguyen and Alexander Mattick},
      year={2023},
      eprint={2304.07327},
      archivePrefix={arXiv},
      primaryClass={cs.CL}
}

@misc{wang2023selfinstruct,
      title={Self-Instruct: Aligning Language Models with Self-Generated Instructions}, 
      author={Yizhong Wang and Yeganeh Kordi and Swaroop Mishra and Alisa Liu and Noah A. Smith and Daniel Khashabi and Hannaneh Hajishirzi},
      year={2023},
      eprint={2212.10560},
      archivePrefix={arXiv},
      primaryClass={cs.CL}
}

@misc{zhou2023lima,
      title={LIMA: Less Is More for Alignment}, 
      author={Chunting Zhou and Pengfei Liu and Puxin Xu and Srini Iyer and Jiao Sun and Yuning Mao and Xuezhe Ma and Avia Efrat and Ping Yu and Lili Yu and Susan Zhang and Gargi Ghosh and Mike Lewis and Luke Zettlemoyer and Omer Levy},
      year={2023},
      eprint={2305.11206},
      archivePrefix={arXiv},
      primaryClass={cs.CL}
}

@misc{chen2023maybe,
      title={Maybe Only 0.5\% Data is Needed: A Preliminary Exploration of Low Training Data Instruction Tuning}, 
      author={Hao Chen and Yiming Zhang and Qi Zhang and Hantao Yang and Xiaomeng Hu and Xuetao Ma and Yifan Yanggong and Junbo Zhao},
      year={2023},
      eprint={2305.09246},
      archivePrefix={arXiv},
      primaryClass={cs.AI}
}

@misc{alpaca,
  author = {Rohan Taori and Ishaan Gulrajani and Tianyi Zhang and Yann Dubois and Xuechen Li and Carlos Guestrin and Percy Liang and Tatsunori B. Hashimoto },
  title = {Stanford Alpaca: An Instruction-following LLaMA model},
  year = {2023},
  publisher = {GitHub},
  journal = {GitHub repository},
  howpublished = {\url{https://github.com/tatsu-lab/stanford_alpaca}},
}

@misc{gpt4all,
  author = {Yuvanesh Anand and Zach Nussbaum and Brandon Duderstadt and Benjamin Schmidt and Andriy Mulyar},
  title = {GPT4All: Training an Assistant-style Chatbot with Large Scale Data Distillation from GPT-3.5-Turbo},
  year = {2023},
  publisher = {GitHub},
  journal = {GitHub repository},
  howpublished = {\url{https://github.com/nomic-ai/gpt4all}},
}

@misc{touvron2023llama,
      title={LLaMA: Open and Efficient Foundation Language Models}, 
      author={Hugo Touvron and Thibaut Lavril and Gautier Izacard and Xavier Martinet and Marie-Anne Lachaux and Timothée Lacroix and Baptiste Rozière and Naman Goyal and Eric Hambro and Faisal Azhar and Aurelien Rodriguez and Armand Joulin and Edouard Grave and Guillaume Lample},
      year={2023},
      eprint={2302.13971},
      archivePrefix={arXiv},
      primaryClass={cs.CL}
}

@misc{ouyang2022training,
      title={Training language models to follow instructions with human feedback}, 
      author={Long Ouyang and Jeff Wu and Xu Jiang and Diogo Almeida and Carroll L. Wainwright and Pamela Mishkin and Chong Zhang and Sandhini Agarwal and Katarina Slama and Alex Ray and John Schulman and Jacob Hilton and Fraser Kelton and Luke Miller and Maddie Simens and Amanda Askell and Peter Welinder and Paul Christiano and Jan Leike and Ryan Lowe},
      year={2022},
      eprint={2203.02155},
      archivePrefix={arXiv},
      primaryClass={cs.CL}
}

@misc{gudibande2023false,
      title={The False Promise of Imitating Proprietary LLMs}, 
      author={Arnav Gudibande and Eric Wallace and Charlie Snell and Xinyang Geng and Hao Liu and Pieter Abbeel and Sergey Levine and Dawn Song},
      year={2023},
      eprint={2305.15717},
      archivePrefix={arXiv},
      primaryClass={cs.CL}
}

@misc{brown2020language,
      title={Language Models are Few-Shot Learners}, 
      author={Tom B. Brown and Benjamin Mann and Nick Ryder and Melanie Subbiah and Jared Kaplan and Prafulla Dhariwal and Arvind Neelakantan and Pranav Shyam and Girish Sastry and Amanda Askell and Sandhini Agarwal and Ariel Herbert-Voss and Gretchen Krueger and Tom Henighan and Rewon Child and Aditya Ramesh and Daniel M. Ziegler and Jeffrey Wu and Clemens Winter and Christopher Hesse and Mark Chen and Eric Sigler and Mateusz Litwin and Scott Gray and Benjamin Chess and Jack Clark and Christopher Berner and Sam McCandlish and Alec Radford and Ilya Sutskever and Dario Amodei},
      year={2020},
      eprint={2005.14165},
      archivePrefix={arXiv},
      primaryClass={cs.CL}
}

@article{UnsupervisedMultitaskLearners,
  author={Radford, Alec and Wu, Jeffrey and Child, Rewon and Luan, David and Amodei, Dario and Sutskever, Ilya},
  description = {Language Models are Unsupervised Multitask Learners},
  title = {Language Models are Unsupervised Multitask Learners},
  url = {https://d4mucfpksywv.cloudfront.net/better-language-models/language-models.pdf},
  year = 2018
}

@misc{schulman2017proximal,
      title={Proximal Policy Optimization Algorithms}, 
      author={John Schulman and Filip Wolski and Prafulla Dhariwal and Alec Radford and Oleg Klimov},
      year={2017},
      eprint={1707.06347},
      archivePrefix={arXiv},
      primaryClass={cs.LG}
}

@misc{liang2022holistic,
      title={Holistic Evaluation of Language Models}, 
      author={Percy Liang and Rishi Bommasani and Tony Lee and Dimitris Tsipras and Dilara Soylu and Michihiro Yasunaga and Yian Zhang and Deepak Narayanan and Yuhuai Wu and Ananya Kumar and Benjamin Newman and Binhang Yuan and Bobby Yan and Ce Zhang and Christian Cosgrove and Christopher D. Manning and Christopher Ré and Diana Acosta-Navas and Drew A. Hudson and Eric Zelikman and Esin Durmus and Faisal Ladhak and Frieda Rong and Hongyu Ren and Huaxiu Yao and Jue Wang and Keshav Santhanam and Laurel Orr and Lucia Zheng and Mert Yuksekgonul and Mirac Suzgun and Nathan Kim and Neel Guha and Niladri Chatterji and Omar Khattab and Peter Henderson and Qian Huang and Ryan Chi and Sang Michael Xie and Shibani Santurkar and Surya Ganguli and Tatsunori Hashimoto and Thomas Icard and Tianyi Zhang and Vishrav Chaudhary and William Wang and Xuechen Li and Yifan Mai and Yuhui Zhang and Yuta Koreeda},
      year={2022},
      eprint={2211.09110},
      archivePrefix={arXiv},
      primaryClass={cs.CL}
}

@article{47761,
title	= {Natural Questions: a Benchmark for Question Answering Research},
author	= {Tom Kwiatkowski and Jennimaria Palomaki and Olivia Redfield and Michael Collins and Ankur Parikh and Chris Alberti and Danielle Epstein and Illia Polosukhin and Matthew Kelcey and Jacob Devlin and Kenton Lee and Kristina N. Toutanova and Llion Jones and Ming-Wei Chang and Andrew Dai and Jakob Uszkoreit and Quoc Le and Slav Petrov},
year	= {2019},
journal	= {Transactions of the Association of Computational Linguistics}
}

@misc{longpre2023flan,
      title={The Flan Collection: Designing Data and Methods for Effective Instruction Tuning}, 
      author={Shayne Longpre and Le Hou and Tu Vu and Albert Webson and Hyung Won Chung and Yi Tay and Denny Zhou and Quoc V. Le and Barret Zoph and Jason Wei and Adam Roberts},
      year={2023},
      eprint={2301.13688},
      archivePrefix={arXiv},
      primaryClass={cs.AI}
}

@misc{OpenLLMLeaderboard,   
    title = {Open LLM Leaderboard},   
    url = {https://huggingface.co/spaces/HuggingFaceH4/open_llm_leaderboard},   
    author = {Huggingface},
    year = {2023},   
    note = {Accessed: 2023-06-10}
}

@software{eval-harness,
  author       = {Gao, Leo and
                  Tow, Jonathan and
                  Biderman, Stella and
                  Black, Sid and
                  DiPofi, Anthony and
                  Foster, Charles and
                  Golding, Laurence and
                  Hsu, Jeffrey and
                  McDonell, Kyle and
                  Muennighoff, Niklas and
                  Phang, Jason and
                  Reynolds, Laria and
                  Tang, Eric and
                  Thite, Anish and
                  Wang, Ben and
                  Wang, Kevin and
                  Zou, Andy},
  title        = {A framework for few-shot language model evaluation},
  month        = sep,
  year         = 2021,
  publisher    = {Zenodo},
  version      = {v0.0.1},
  doi          = {10.5281/zenodo.5371628},
  url          = {https://doi.org/10.5281/zenodo.5371628}
}

@misc{kumar2022finetuning,
      title={Fine-Tuning can Distort Pretrained Features and Underperform Out-of-Distribution}, 
      author={Ananya Kumar and Aditi Raghunathan and Robbie Jones and Tengyu Ma and Percy Liang},
      year={2022},
      eprint={2202.10054},
      archivePrefix={arXiv},
      primaryClass={cs.LG}
}

@misc{ChatGPT,
    title={ChatGPT: Optimizing language models for dialogue.},
    author={OpenAI},
    url={https://online-chatgpt.com/},
    year={2022}
}

@misc{Bard,
    author={Sundar Pichai},
    title={An important next step on our AI journey. Google AI Blog},
    url={https://blog.google/intl/en-africa/products/explore-get-answers/an-important-next-step-on-our-ai-journey/},
    year={2023}
}

@misc{Claude,
    author={AnthropicAI},
    title={Introducing Claude},
    url={https://www.anthropic.com/index/introducing-claude},
    year={2023}
}

@misc{hinton2015distilling,
      title={Distilling the Knowledge in a Neural Network}, 
      author={Geoffrey Hinton and Oriol Vinyals and Jeff Dean},
      year={2015},
      eprint={1503.02531},
      archivePrefix={arXiv},
      primaryClass={stat.ML}
}

@misc{zhang2022opt,
      title={OPT: Open Pre-trained Transformer Language Models}, 
      author={Susan Zhang and Stephen Roller and Naman Goyal and Mikel Artetxe and Moya Chen and Shuohui Chen and Christopher Dewan and Mona Diab and Xian Li and Xi Victoria Lin and Todor Mihaylov and Myle Ott and Sam Shleifer and Kurt Shuster and Daniel Simig and Punit Singh Koura and Anjali Sridhar and Tianlu Wang and Luke Zettlemoyer},
      year={2022},
      eprint={2205.01068},
      archivePrefix={arXiv},
      primaryClass={cs.CL}
}

@misc{Palmyra,
    author={Writer},
    title={Palmyra LLMs empower secure, enterprise-grade generative AI for business. Writer Blog},
    url={https://writer.com/blog/palmyra/},
    year={2023}
}

@misc{HuggingfaceAutoTrain,
    author={Huggingface},
    title={AutoTrain: Create powerful AI models without code},
    url={https://huggingface.co/autotrain},
    year={2023}
}

@article{gao2020pile,
  title={The Pile: An 800GB Dataset of Diverse Text for Language Modeling},
  author={Gao, Leo and Biderman, Stella and Black, Sid and Golding, Laurence and Hoppe, Travis and Foster, Charles and Phang, Jason and He, Horace and Thite, Anish and Nabeshima, Noa and others},
  journal={arXiv preprint arXiv:2101.00027},
  year={2020}
}

@misc{hendrycks2021measuring,
      title={Measuring Massive Multitask Language Understanding}, 
      author={Dan Hendrycks and Collin Burns and Steven Basart and Andy Zou and Mantas Mazeika and Dawn Song and Jacob Steinhardt},
      year={2021},
      eprint={2009.03300},
      archivePrefix={arXiv},
      primaryClass={cs.CY}
}

@misc{wei2022emergent,
      title={Emergent Abilities of Large Language Models}, 
      author={Jason Wei and Yi Tay and Rishi Bommasani and Colin Raffel and Barret Zoph and Sebastian Borgeaud and Dani Yogatama and Maarten Bosma and Denny Zhou and Donald Metzler and Ed H. Chi and Tatsunori Hashimoto and Oriol Vinyals and Percy Liang and Jeff Dean and William Fedus},
      year={2022},
      eprint={2206.07682},
      archivePrefix={arXiv},
      primaryClass={cs.CL}
}
\end{document}